\documentclass{bmvc2k}

\usepackage{tikz}
\usepackage{comment}
\usepackage{amsmath,amssymb} 
\usepackage{color}
\usepackage{booktabs}
\usepackage{boldline}
\usepackage{colortbl}
\usepackage{booktabs}
\usepackage{bbm}
\usepackage{multirow}
\usepackage{multicol}
\usepackage{makecell}
\usepackage{xcolor}
\usepackage{xspace}
\usepackage{float}

\title{Multiple Object Tracking from appearance by hierarchically clustering tracklets}

\addauthor{Andreu Girbau}{agirbau@nii.ac.jp}{1}
\addauthor{Ferran Marqués}{ferran.marques@upc.edu}{2}
\addauthor{Shin'ichi Satoh}{satoh@nii.ac.jp}{1}

\addinstitution{
 National Institute of Informatics\\
 Tokyo, Japan
}
\addinstitution{
 Universitat Politècnica de Catalunya \\
 Barcelona, Spain
}

\runninghead{Girbau, Marqués, Satoh}{MOT from appearance}

\begin{document}

\maketitle
\begin{abstract}

Current approaches in Multiple Object Tracking (MOT) rely on the spatio-temporal coherence between detections combined with object appearance to match objects from consecutive frames.
In this work, we explore MOT using object appearances as the main source of association between objects in a video, using spatial and temporal priors as weighting factors.
We form initial tracklets by leveraging on the idea that instances of an object that are close in time should be similar in appearance, and build the final object tracks by fusing the tracklets in a hierarchical fashion.
We conduct extensive experiments that show the effectiveness of our method over three different MOT benchmarks, MOT17, MOT20, and DanceTrack, being competitive in MOT17 and MOT20 and establishing state-of-the-art results in DanceTrack.
Code is available at \href{https://github.com/NII-Satoh-Lab/MOT_FCG}{https://github.com/NII-Satoh-Lab/MOT\_FCG}.

\end{abstract}
\section{Introduction}

The objective of the Multiple Object Tracking (MOT) task is to estimate the trajectory of a set of objects (e.g. pedestrians or vehicles) along a video sequence. The objects of interest must be represented with an accurate bounding box, and keep the associated identity over time. MOT can be useful in many applications, such as autonomous driving, robotics, or automatic production of events.

The dominant approach in MOT is tracking-by-detection, a two-step process where, first, the objects of interest are detected frame-by-frame, and then associated across frames to form tracks. This approach became feasible thanks to the latest advances in object detection \cite{ren2015faster,liu2016ssd,redmon2017yolo9000,duan2019centernet,yolox2021}, leading to the current dominant trackers, which use spatio-temporal priors combined with appearance features as the main source of association.

In this paper, we explore the association of objects by primarily focusing on their appearance, using spatial and temporal priors as a support to weight object relationships, and present a simple and adaptable MOT method based on tracklet generation and clustering.
Our tracker, FCG (Feature Combinatorial Grouping), leverages on the idea that a specific object is prone to have a similar appearance in a temporal neighborhood. Based on this, we form high quality tracklets, smaller parts of an object track, for each object in a local neighborhood, capturing small variations of the object appearance, and cluster the generated tracklets in a hierarchical way to form the final tracks, taking into account the sequential nature of a video, without any kind of post-processing or optimization when performing the tracking.

We conduct extensive experiments to show the validity of our method, and verify the quality of FCG in three MOT benchmarks, MOT17 \cite{MOT16}, MOT20 \cite{dendorfer2020mot20}, and DanceTrack \cite{peize2021dance}, achieving comparable performance against current state-of-the art trackers for MOT17 and MOT20, and setting state-of-the-art results for DanceTrack.

Our contributions in this paper are:
\begin{itemize}
    \item We explore the effectiveness of focusing on appearance as the main source for associating objects through time for MOT, and present FCG, a tracker that relies on object appearance features to form tracks by hierarchically clustering tracklets.
    \item We show that a fairly simple tracklet association by clustering method, using off-the-shelf pre-trained models and without any kind of optimization or post-processing, can perform competitively to current, way more complex, state-of-the-art techniques.
    \item We achieve state-of-the-art results for the DanceTrack benchmark, while presenting competitive results in MOT17 and MOT20. 
\end{itemize}

\section{Related work}
Advances in object detection \cite{ren2015faster,redmon2017yolo9000,duan2019centernet,yolox2021} allowed current tracking methods to rely on frame-by-frame detections to perform tracking of multiple objects. This led to the current predominant approach in MOT, tracking-by-detection, which addresses the tracking problem in two steps: (i) detection of the objects in the scene, and (ii) the association of the detections through time to form tracks. 
Simultaneously, object appearance features for Re-Identification have been highly improved, and have been proven very effective \cite{hermans2017defense,ristani2018features,wang2018learning,luo2019bag}.

\subsection{Object appearance in MOT}
Despite some state-of-the-art MOT methods do not use the object appearance at all \cite{bewley2016simple, chen2018real, stadler2022modelling, zhang2022bytetrack}, relying on motion models to associate close objects between frames, object appearance is extensively used in MOT. Some MOT methods account indirectly for object appearance \cite{zhou2020tracking, tokmakov2021learning, zeng2021motr}, where the motion of objects is inferred using previous detections and image features, train models to both learn to detect and represent objects at the same time with multiple output heads \cite{zhang2021fairmot,yu2021relationtrack}, using recurrent neural networks \cite{wang2021general}, attention mechanisms \cite{chen2019aggregate} or correlation operations between convolutional feature maps \cite{wang2021multiple,feichtenhofer2017detect} to encode both motion and feature information, or even with end-to-end approaches where feature extraction, affinity estimation, and objects association are refined in a single network \cite{chu2019famnet}. 

Explicit usage of object appearance is also extensively used in MOT, combining the spatio-temporal information with object appearance information as a weighting factor \cite{wojke2017simple}, optimizing a graph where the position and feature embedding information form the nodes and edges \cite{hornakova2020lifted,braso2020learning}, or to recover lost tracks by doing re-identification \cite{bergmann2019tracking, meinhardt2022trackformer}.

The key difference between our proposed method and previous works is what is prioritized for doing object association, as previous works prioritize spatio-temporal priors or their combination with object appearance to associate objects. In this work, we explore using appearance features to associate the objects as the main source of association, using spatial and temporal information as a support weighting factor. 

\subsection{Tracking by tracklet association}
Tracklets, introduced in \cite{berclaz2006robust}, are smaller parts of whole object tracks that, once fused, form the whole object track.  
Tracking by associating tracklets is a well established technique \cite{huang2008robust}, that has had far less attention than other offline tracking techniques such as graph optimization considering individual detections.
One common assumption in tracking by associating tracklets is to consider tracklets as a robust temporal representation of the object. Once the initial tracklets are built, they are associated to form the final tracks. 

Some methods associate tracklets by using a MAP (Maximum A Posteriory) approach, conditioning the overall result every time a tracklet is fused \cite{huang2008robust}, cluster tracklets based on color, spatial and temporal attributes \cite{topkaya2016tracklet}, form tracks by clustering objects using spatial cues \cite{zhang2017multi, stadler2021multi}, gradually add similar objects to the initial tracklets \cite{wu2018exploit}, or treat tracklets as the nodes of a graph, and the similarity between tracklets as the vertexes \cite{ma2018customized,wang2019exploit,zhang2020long,dai2021learning}.

Tracklet association methods are usually treated as an optimization problem, whether it is iteratively updating cluster assignments \cite{topkaya2016tracklet,wu2018exploit,stadler2021multi}, using a graph formulation with the subsequent optimization \cite{zhang2017multi,ma2018customized,wang2019exploit,zhang2020long,dai2021learning}, or re-training CNN networks to adapt ReID features to new sequences \cite{ma2018customized}.
Our method hierarchically fuses tracklets based on the objects appearance considering the sequential nature of video object tracking \textit{without requiring any type of optimization or post-processing} within the tracking pipeline.

\vspace{-1mm}
\section{Method}
\vspace{-1mm}

\begin{figure}[t]
    \centering
    \includegraphics[width=\textwidth]{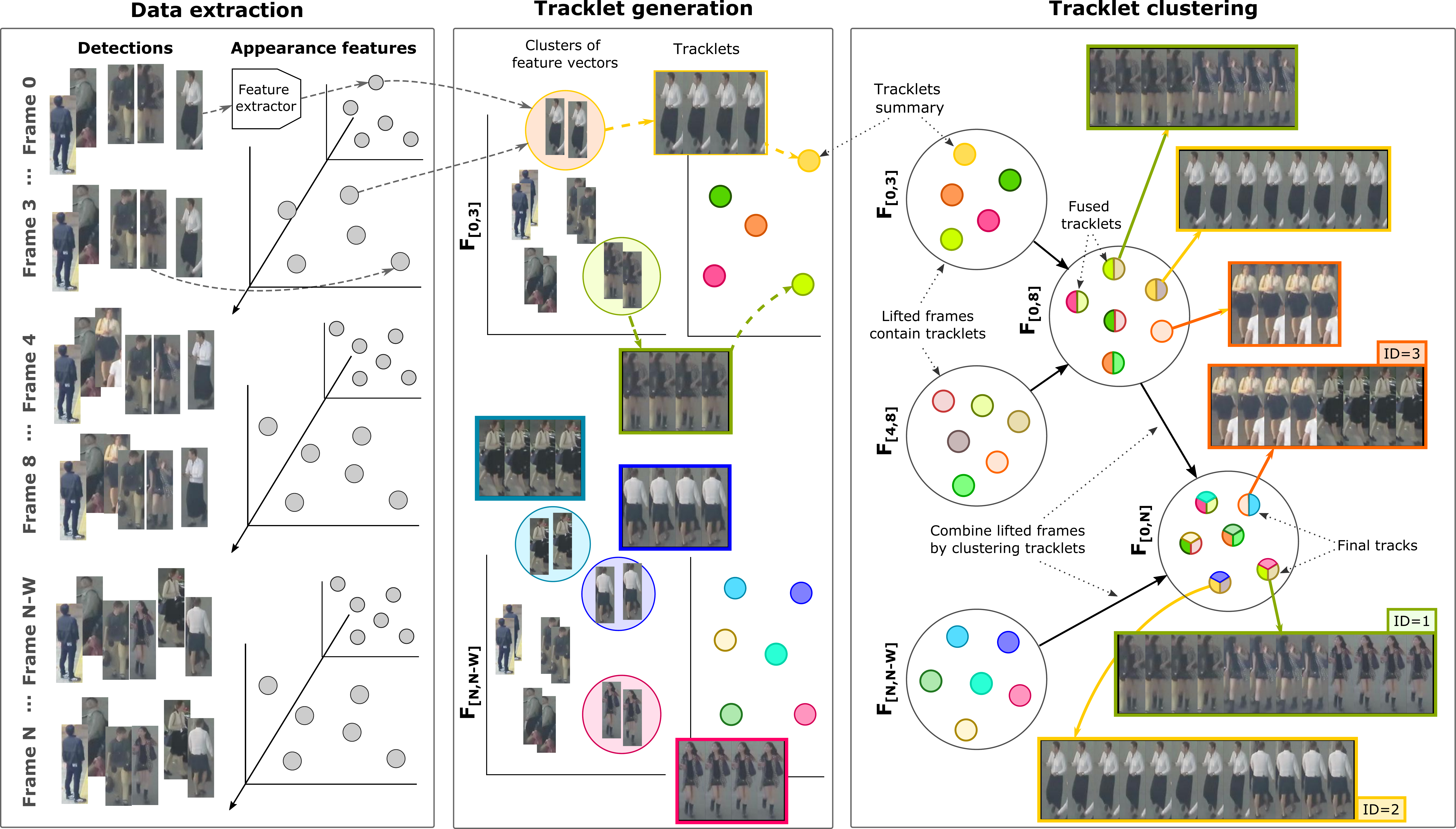}
    \caption{We track objects by clustering their appearance features. We build tracklets to have a robust representation of the objects in a temporal neighborhood, while capturing small variations of the object appearance, and hierarchically cluster the tracklets to form the final tracks.}
    \label{fig:fig_teaser}
\end{figure}

FCG leverages on the idea that instances of the same object have similar appearance in a temporal neighborhood. It consists on two stages: the first stage generates an initial set of short, but reliable, tracklets, and the second stage fuses these tracklets over time by clustering them in a sequentially-guided hierarchical way, naturally leading to the final object tracks. 
To generate tracklets and further fuse them, we use the implementation \cite{2020SciPy-NMeth} of UPGMA (Unweighted Pair Group Method with Arithmetic mean) \cite{sneath1973unweighted}, which iteratively fuses pairs of clusters, forming a hierarchy. In Figure \ref{fig:fig_teaser} we show FCG pipeline.

\smallskip
\noindent\textbf{Tracklets}

A tracklet $T$ is defined as a set of instances of a specific object being tracked between two time instants. The union of the object tracklets through a whole video sequence will form the final object track.
To form the initial set of tracklets, we cluster the object instances present in the frames contained in a non-overlapping temporal window $W$ of size $\tau$. For a video sequence, we will have $N$ possible temporal windows.

\smallskip
\noindent\textbf{Lifted frames}

To hierarchically cluster tracklets, we introduce the concept of \textit{lifted frames}, consisting on artificial time instants that contain tracklets instead of individual, discrete, detections. 
Formally, we define a lifted frame $F^{i}_{[n, m]}$ as the set of tracklets that contain detections from time $t=n \cdot \tau$ to $t=m \cdot \tau$, where $i \in [0, N] \in \mathbb{N}$ corresponds the current level of the hierarchy, and $n \in [0, N-1]$ and $m \in [1, N], \{n,m \in \mathbb{N} | m > n\}$ correspond to the lifted frame temporal indexes. Note that each tracklet $T^{k} = F^{i,k}_{[n, m]}$ present in a lifted frame $F^{i}_{[n, m]}$ has a length $l \in [1, (m-n)\tau]$.

\subsection{Stage zero: extracting the data}
As a pre-processing step we first detect all the objects in a video using a given object detector, in this work YOLOX (You Only Look Once X) \cite{yolox2021}, and extract their appearance features with a feature-extractor network, in this case SBS (Stronger Baseline of Specials) \cite{he2020fastreid}. 

\subsection{First stage: tracklet generation}
Once we have all the detections in each frame and their associated appearance feature vectors, we proceed to generate the initial set of tracklets for the objects of interest.
These tracklets will be a representation of an object in a temporal neighborhood, capturing small variations of the object appearance that will be key to form the final tracks.

To compare two object instances, $O^k$ and $O^p$, we use the cosine distance over the corresponding appearance features, $\bf h^k$ and $\bf h^p$ $\in \mathbb{R}^{2048}$. 

\begin{equation}
    d(O^k, O^p) = 1 - \frac{{\bf h^{k\top}} {\bf h^p}}{\|{\bf h^k}\| \|{\bf h^p}\|}
\end{equation}

To generate tracklets, we cluster object instances within a temporal window $W$ of size $\tau$.
We constrain the fusion of objects to different frames, i.e., two objects from the same frame will not be able to be clustered in the same tracklet. In this work, we use $W=6$, and set the cluster fusion threshold to $0.055$.

\subsection{Second stage: tracklet clustering}
Once the object tracklets are formed for each lifted frame $F^{1}_{[n, n+1]}$, for $n \in [0, N-1]$, we fuse them in a hierarchical way. Whenever a set of lifted frames are fused, i.e., the tracklets in the lifted frames are clustered together, the current level of the hierarchy $i$ is increased by $1$. 

To efficiently compare two tracklets, $T^k=F^{i,k}_{[n, m]}$ and $T^q=F^{i,q}_{[u, v]}$, we summarize the collection of appearance feature vectors present in each tracklet, $\mathcal{H}^k$, $\mathcal{H}^q$, as their element-wise median, and compute the cosine distance between both.

\begin{equation}
    d(T^k, T^q) = 1 - \frac{{\bf \Tilde{h}^{k\top}} {\bf \Tilde{h}^q}}{\|{\bf \Tilde{h}^k}\| \|{\bf \Tilde{h}^q}\|} 
    \ , \
    \bf \Tilde{h}^k = \text{Med}(\mathcal{H}^{k})
    \ , \
    \bf \Tilde{h}^q = \text{Med}(\mathcal{H}^{q})
\end{equation}

Generically, we define the fusion of two lifted frames as the clustering of the union of the tracklets contained in each lifted frame.

\begin{equation}
    F^{i+1}_{[n, m] \cup [u, v]} = F^{i}_{[n, m]} \cup F^{i}_{[u, v]}
\end{equation}

For the MOT setting, we constrain the fusion to consecutive lifted frames, following the timeline of the video sequence. This is:

\begin{equation}
    F^{i+1}_{[n, u]} = F^{i}_{[n, m]} \cup F^{i}_{[m+1, u]} \ , \  \{n<m<u\}
\end{equation}

\noindent In the ablation study of Sec. \ref{sec:4_analysis} we further analyse the impact of this weighting.

By iteratively clustering tracklets by fusing lifted frames, we naturally get to a point where all tracklets are defined within a single lifted frame. This happens at $F^{N}_{[0, N]}$, where the clustering of tracklets is at the top level $N$ of the hierarchy. We assign an ID to each tracklet, becoming the final track for each tracked object.

\subsection{Multiple object tracking}
\label{sec:3_mot}

We use FCG for the MOT task given the natural way the clustered tracklets form the final tracks of objects. Furthermore, we consider some additions with regards to the MOT challenge that can be embedded into the clustering scheme.

\smallskip
\textbf{Temporal coherence.} 
It is a common practice in MOT trackers to terminate the tracking of an object if no new detections are associated to that object within a specified time.

We incorporate a similar idea to FCG so that, if two tracklets $T^k$ and $T^q$ are further than a specified temporal distance $K_{T}$, given the last detection in $T^k$, $B^k$, and the first detection in $T^q$, $B^q$, the fusion between the nodes of the hierarchy will be harder, but not impossible, as FCG relies on appearance features to associate tracks, being able to recover from longer-term occlusions. We achieve this by multiplying the distance between two tracklets that are further than $K_{T}$ by a constant $\lambda_{T} \geq 1$, making the two tracklets $\lambda_{T}$ times harder to be associated, only allowing very similar tracklets to be fused.

\begin{equation}
    d'(T^k, T^q) = d(T^k, T^q) \cdot \lambda_{T}, \quad
    \lambda_{T}=\begin{cases}
    1 & \Delta{t}(B^k,B^q) \leq K_{T}\\
    c_T & \Delta{t}(B^k,B^q) > K_{T}
    \end{cases}
\end{equation}

In this work, we set the constant value to $c_T=4$, and $K_T=40$.

\textbf{Spatial coherence.}
As many objects share similar features, we also take into account spatial information, making closer objects easier to fuse, while constraining the fusion of farther objects.

For two tracklets $T^k$ and $T^q$, we compute both the IoU (Intersection over Union) and a L2 distance normalized w.r.t. bounding box size between the last detection in $T^k$, $B^k$, and the first detection in $T^q$, $B^q$. We multiply the original distance between tracklets $d(T^k, T^q)$ by $\lambda_C$ and $\lambda_F$, corresponding to close and far objects respectively.

\begin{equation}
    d'(T^k, T^q) = d(T^k, T^q) \cdot \lambda_{C} \cdot \lambda_{F}
\end{equation}
\[
    \lambda_{C}=\min(1, d_{\text{IoU}} + \text{off}), \quad
    \lambda_{F}=\begin{cases}
    1 & d_{\text{box}}(B^k, B^q) \leq K_{F}\\
    c_F & d_{\text{box}}(B^k, B^q) > K_{F}
    \end{cases}
\]

\noindent We set $\text{off}=0.15$ to avoid multiplying by 0 the distance when two objects are completely overlapped, $c_F=2$ to make spatially far detections twice as hard to be fused, and $K_{F}=2$ as the normalized distance between bounding boxes. Note that $\lambda_{C} \in [\text{off}, 1]$, easing the fusion between tracklets, and $\lambda_{C} \in \{1,c_F\}$, making them harder to fuse, but not impossible. We convert the IoU between $B^k$ and $B^q$ to the distance $d_{\text{IoU}}$,

\begin{equation}
    d_{\text{IoU}}(B^k, B^q) = 1 - \frac{B^k \cap B^q}{B^k \cup B^q}
\end{equation}

\noindent and compute the normalised displacement between $B^k$ and $B^q$ as the distance $d_{\text{box}}$ between their extreme points (left, top, right, bottom),

\begin{equation}
    d_{\text{box}}(B^k, B^q) = \frac{d_1 + d_2}{2}
\end{equation}

\noindent where $d_1$ corresponds to the distance between the left-top corners of the bounding boxes $B^k$ and $B^q$, and $d_2$ to the distance between the right-bottom corners of the bounding boxes $B^k$ and $B^q$. Both distances are normalized by the average of the bounding boxes width and height to have in consideration the size of the bounding box when computing the displacement.

\[
    d_1 = \sqrt[]{\left(\frac{B^k_{\text{left}} - B^q_{\text{left}}}{\frac{B^k_{\text{width}} + B^q_{\text{width}}}{2}} \right)^2 + \left(\frac{B^k_{\text{top}} - B^q_{\text{top}}}{\frac{B^k_{\text{height}} + B^q_{\text{height}}}{2}} \right)^2} 
\]

\[
    d_2 = \sqrt[]{\left(\frac{B^k_{\text{right}} - B^q_{\text{right}}}{\frac{B^k_{\text{width}} + B^q_{\text{width}}}{2}} \right)^2 + \left(\frac{B^k_{\text{bot}} - B^q_{\text{bot}}}{\frac{B^k_{\text{height}} + B^q_{\text{height}}}{2}} \right)^2}
\]

With this, the distance between tracklets $T^k$ and $T^q$ can be weighted by both temporal and spatial priors. 

\begin{equation}
    d'(T^k, T^q) = d(T^k, T^q) \cdot \lambda_{T} \cdot \lambda_{C} \cdot \lambda_{F}
\end{equation}

\textbf{Motion estimation.}
We integrate a simple motion estimation, consisting on a constant velocity assumption, to help in the spatial coherence. This motion is calculated as the difference between the previous $t-1$ and current $t$ positions of a bounding box $B$ existing in a tracklet, and applied to displace $B$ to the next time step $t+1$. 
\section{Experimental evaluation}

\noindent\textbf{Datasets.}
Experiments are performed in three different MOT datasets: MOT17 \cite{MOT16}, MOT20 \cite{dendorfer2020mot20}, and DanceTrack \cite{peize2021dance}.
MOT17 and MOT20 address the problem of tracking pedestrians in the wild, while DanceTrack focuses on a group of dancers on a stage.
Following the work in MOT17, MOT20 has a much larger density of pedestrians in the scene, longer sequences, and the camera is set further away. 
In DanceTrack, where the setting is much more artistic, the main difficulties consist on the amount of very similarly dressed dancers, appearance changes due to zooms or moving cameras, the amount of occlusions and grouping, and the diverse motion, which, in contrast with MOT17 and MOT20, can be much more random.

\begin{figure}[h]
    \centering
    \includegraphics[width=0.85\textwidth]{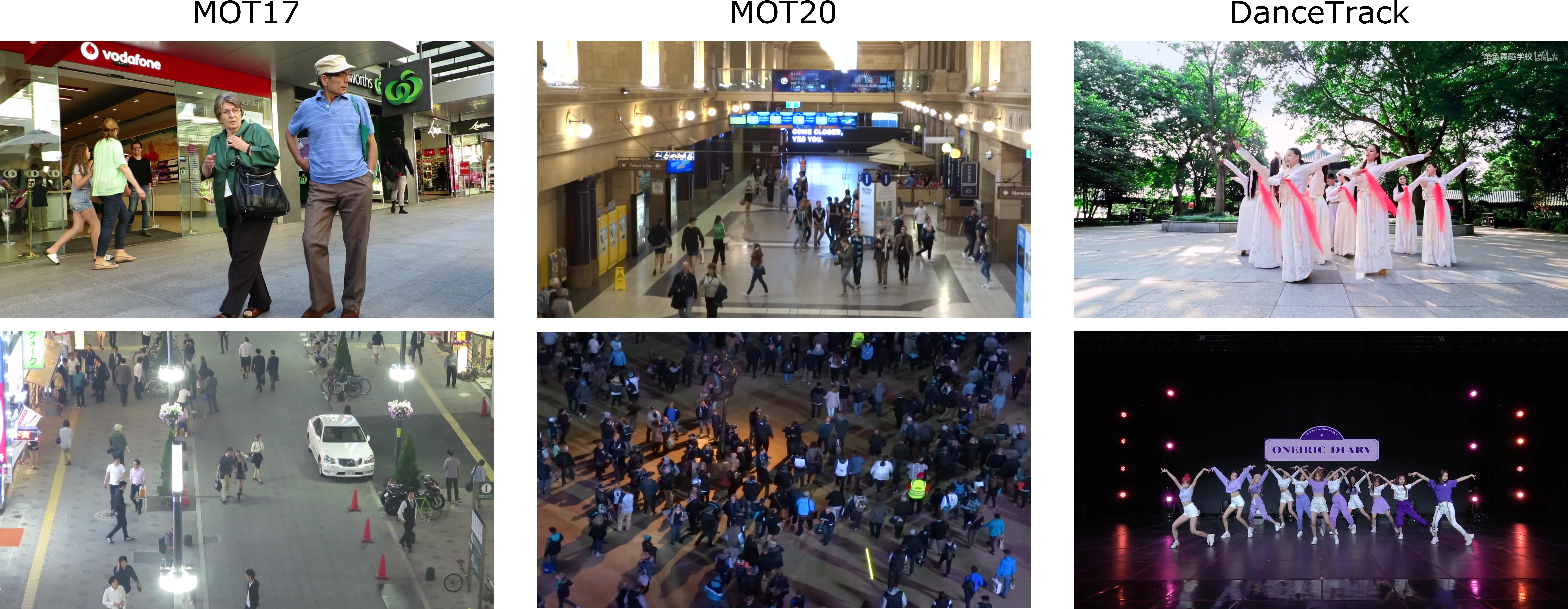}
    \caption{We test FCG on three different benchmarks. MOT17 and MOT20 focus on tracking pedestrians in the wild, and DanceTrack on tracking dancers on a stage.}
    \label{fig:datasets}
\end{figure}

\noindent\textbf{Metrics.}
To evaluate our method, we use the main established metrics for MOT, which are HOTA \cite{luiten2020hota}, built as the combination of DetA (Detection Accuracy score) and AssA (Association Accuracy score), MOTA \cite{bernardin2008evaluating}, and IDF1 \cite{ristani2016performance}. For all metrics, higher is better.

\smallskip
\noindent\textbf{Implementation details.}
To detect the objects in the scene we use YOLOX \cite{yolox2021}, pre-trained with COCO \cite{lin2014microsoft} and fine-tuned on each dataset. In this work, we use a detector threshold of $0.7$. For the extraction of appearance features we use SBS \cite{he2020fastreid} trained with Market1501 \cite{zheng2015scalable}. We conducted experiments using different detectors and feature extraction networks in Sec. \ref{sec:4_analysis}.
For all benchmarks, we build the initial tracklets from detections within a sliding window of $6$ frames.

If not stated otherwise, for MOT17 and MOT20, where pedestrians may leave the scene to never reappear, we use the temporal constraint and motion estimation introduced in Sec. \ref{sec:3_mot}. For DanceTrack, where is usual for dancers to reappear in the scene, we don't use neither the temporal constraint nor motion estimation. For all benchmarks, we use the spatial coherence constraints, also introduced in Sec. \ref{sec:3_mot}, where two tracklets are easier or more difficult to fuse depending on the spatial distance between their corresponding detections.




\subsection{Comparison to the state-of-the-art}
In Table \ref{tab:sota}, we report results of the performance of FCG on the test set of each respective benchmark, and compare it to other state-of-the-art methods. All method use their own, private, detectors. The results for each dataset are presented as the overall results for all sequences. 

By comparing to other state-of-the-art results, FCG shows that similar tracking performance can be achieved by using appearance between objects as the basic source of information to associate objects over time, with a much simpler technique and without the need of any kind of optimization or post-processing when performing the tracking. 
This raises the intuition that having strong object information, in this case detections and person re-identification features, can be as important as the tracking method itself.

Overall, FCG presents competitive results compared to other methods on MOT17 and MOT20 datasets, achieving state-of-the-art results for the DanceTrack benchmark.

{\begin{table}
\centering

\setlength{\tabcolsep}{0.6mm}

\resizebox{1\linewidth}{!}{
\begin{tabular}{l | ccccc | ccccc | ccccc}

\arrayrulecolor{white}\hline
\Xhline{2\arrayrulewidth}
\arrayrulecolor{white}\hline

\arrayrulecolor{black}
\multirow{2}{*}{Methods} & \multicolumn{5}{c|}{MOT17~\cite{MOT16}} & \multicolumn{5}{c|}{MOT20~\cite{dendorfer2020mot20}} & \multicolumn{5}{c}{DanceTrack~\cite{peize2021dance}} \\
& HOTA & DetA & AssA & MOTA & IDF1 &
  HOTA & DetA & AssA & MOTA & IDF1 &
  HOTA & DetA & AssA & MOTA & IDF1\\
\arrayrulecolor{black}\hline
CenterTrack~\cite{zhou2020tracking} & 52.2 & 53.8 & 51.0 & 67.8 & 64.7 & - & - & - & - & - & 41.8 & 78.1 & 22.6 & 86.8 & 35.7\\
FairMOT~\cite{zhang2021fairmot} & 59.3 & 60.9 & 58.0 & 73.7 & 72.3 & 54.6 & 54.7 & 54.7 & 61.8 & 67.3 & 39.7 & 66.7 & 23.8 & 82.2 & 40.8 \\
QDTrack~\cite{pang2021quasi} & 53.9 & 55.6 & 52.7 & 68.7 & 66.3& - & - & - & - & - & 45.7 & 72.1 & 29.2 & 83.0 & 44.8\\
TransTrack~\cite{sun2020transtrack} &  54.1 & 61.6 & 47.9 & 75.2 & 63.5 & 48.9 & 53.3 & 45.2 & 65.0 & 59.4 & 45.5 & 75.9 & 27.5 & 88.4 & 45.2\\
TraDes~\cite{wu2021track} & 52.7 & 55.2 & 50.8 & 69.1 & 63.9 & - & - & - & - & - & 43.3 & 74.5 & 25.4 & 86.2 & 41.2\\

MOTR~\cite{zeng2021motr} & 55.1 & 56.2 & 54.2 & 67.4 & 67.0 & 43.6 & 51.8 & 37.0 & 59.1 & 49.9 & 48.4 & 71.8 & \textbf{32.7} & 79.2 & 46.1\\
Trackformer~\cite{meinhardt2022trackformer} & 57.3 & 60.9 & 54.1 & 74.1 & 68.0 & 54.7 & 56.7 & 53.0 & 68.6 & 65.7 & - & - & - & - & -\\

ReMOT~\cite{yang2021remot} & 59.7 & 62.8 & 57.1 & 77.0 & 72.0 & 61.2 & \textbf{63.9} & 58.7 & 77.4 & 73.1 & - & - & - & - & - \\
CrowdTrack~\cite{stadler2021performance} & 60.3 & 61.5 & 59.3 & 75.6 & 73.6 & 55.0 & 57.7 & 52.6 & 70.7 & 68.2 & - & - & - & - & - \\

MAA~\cite{stadler2022modelling} & 62.0 & 64.2 & 60.2 & 79.4 & 75.9 & 57.3 & 59.7 & 55.1 & 73.9 & 71.2 & - & - & - & - & - \\
BYTE~\cite{zhang2022bytetrack} & \textbf{63.1} & \textbf{64.5} & 62.0 & \textbf{80.3} & 77.3 & \textbf{61.3} & 63.4 & \textbf{59.6} & \textbf{77.8} & \textbf{75.2} & 47.7 & 71.0 & 32.1 & 89.6 & \textbf{53.9}\\
\textbf{FCG (Ours)} & 62.6 & 62.2 & \textbf{63.4} & 76.7 & \textbf{77.7} & 57.3 & 56.7 & 58.1 & 68.0 & 69.7 & \textbf{48.7} & \textbf{79.8} & 29.9 & \textbf{89.9} & 46.5\\
\arrayrulecolor{white}\hline
\Xhline{2\arrayrulewidth}
\arrayrulecolor{white}\hline
\end{tabular}
}
\caption{Comparison between our method, \textit{FCG}, to the state-of-the-art on the test set of MOT17, MOT20, and DanceTrack. Results of all methods over private detections.}
\label{tab:sota}
\end{table}
}

\smallskip
\subsection{Analysis}
\label{sec:4_analysis}

\textbf{Design choices for tracklet generation}

\begin{figure}[h]
    \centering
    \includegraphics[width=\linewidth]{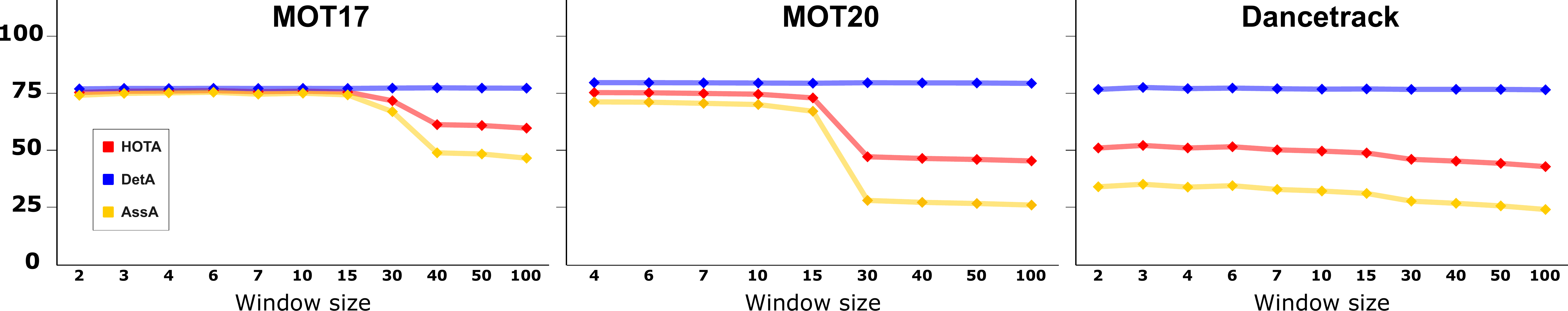}
    \caption{FCG tracking performance for tracklet construction over different window sizes.}
    \label{fig:4_hyperparams}
\end{figure}

In Figure \ref{fig:4_hyperparams} we highlight the effect in performance when choosing different sizes of the temporal window $W$ to build the initial tracklets. 
For all benchmarks, there is a similar performance when forming the tracklets within temporal windows of size $W=2$ to $W=6$ frames, starting to decrease from $W=7$ for the DanceTrack dataset, and from $W=15$ for MOT17 and MOT20. 
This is due to the appearance change in the person being tracked when considering longer tracklets, as many false positives can be added to the tracklet.
For all benchmarks, tracking by directly clustering individual detections was computationally unfeasible. The same applies to window sizes of $W=2$ and $W=3$ in MOT20, where large amount of elements to be tracked are present.

\newpage
\noindent \textbf{Ablation study}

We perform a set of ablations to FCG, considering the enhancements presented in Sec. \ref{sec:3_mot}, and report the results in Table \ref{tab:ablations}. Throughout the ablation, detection-based scores (DetA and MOTA) are very similar. This is due to be using all detections, without adding or discarding any, being the small variations related to ID switching. For this, we only highlight the overall score, HOTA, and the scores related to the association, AssA and IDF1.

First, comparing FCG baselines (first two rows in Table \ref{tab:ablations}), we highlight the fact that following a consecutive (C) order when clustering tracklets is important for MOT17 and MOT20, given the sequential nature of the pedestrian movement, but not for DanceTrack, due to the random dance formation changes, combined with the resemblance of the dancers, producing many ID switches between tracks during transitions.

\begin{table}
\centering

\setlength{\tabcolsep}{1mm}

\resizebox{1\linewidth}{!}{

\begin{tabular}{c c c c |c c c c c | c c c c c | c c c c c}

\arrayrulecolor{white}\hline
\Xhline{2\arrayrulewidth}
\arrayrulecolor{white}\hline

\arrayrulecolor{black}
\multirow{2}{*}{C} &
\multirow{2}{*}{S} & 
\multirow{2}{*}{M} & 
\multirow{2}{*}{T} & 
\multicolumn{5}{c|}{MOT17}  &
\multicolumn{5}{c|}{MOT20}  &
\multicolumn{5}{c}{DanceTrack} \\
& & & & HOTA & DetA & AssA & MOTA & IDF1 &  
  HOTA & DetA & AssA & MOTA & IDF1 &
  HOTA & DetA & AssA & MOTA & IDF1\\
\arrayrulecolor{white}\hline
\arrayrulecolor{black}\hline
\arrayrulecolor{white}\hline

&  &  &  & 61.6 & 77.2 & 49.6 & 86.6 & 64.1 & 33.6 & 79.5 & 14.3 & 91.4 & 27.7 & 46.2 & 77.8 & 27.6 & 82.8 & 41.6\\
\checkmark &  &  &  & 67.6 & 77.2 & 59.6 & 86.7 & 71.9 & 43.3 & 79.5 & 23.7 & 91.8 & 40.1 & 41.8 & 77.9 & 22.5 & 82.8 & 37.0\\
\checkmark & \checkmark & & & 71.9 & 77.3 & 67.2 & 87.3 & 77.4 & 58.4 & 79.8 & 42.9 & 92.4 & 57.0 & \textbf{50.0} & 77.2 & \textbf{32.5} & 87.5 & \textbf{51.8}\\
\checkmark & \checkmark & \checkmark & & 72.4 & 77.5 & 68.0 & 87.3 & 78.2 &    58.2 & 79.8 & 42.4 & 92.4 & 56.2 & 48.9 & 77.2 & 31.1 & 87.4 & 50.8\\
\checkmark & & & \checkmark & 72.3 & 77.2 & 68.0 & 86.7 & 79.0 & 57.6 & 79.5 & 41.9 & 91.8 & 62.5 & 38.9 & 77.8 & 19.6 & 82.9 & 33.4\\
\checkmark & \checkmark & & \checkmark & 72.4 & 77.5 & 68.0 & 87.3 & 78.2 & \textbf{74.8} & 79.8 & \textbf{70.1} & 92.4 & \textbf{85.4} & 48.9 & 77.2 & 31.1 & 87.4 & 50.8\\
\checkmark & \checkmark & \checkmark & \checkmark & \textbf{76.3} & 77.5 & \textbf{75.4} & 87.3 & \textbf{85.1} & 74.6 & 79.8 & 69.8 & 92.4 & 85.2 & 49.1 & 77.3 & 31.4 & 87.5 & 48.9\\
\arrayrulecolor{white}\hline
\Xhline{2\arrayrulewidth}
\arrayrulecolor{white}\hline

\end{tabular}
}
\caption{Ablations on FCG. C stands for Consecutive clustering of tracklets, S for Spatial coherence, M for Motion estimation, and T for Temporal coherence.}
\label{tab:ablations}
\end{table}

Spatial and temporal enhancements do not apply to the non-consecutive baseline due to the need of spatio-temporal coherence when comparing the detections in tracklets. 
To the FCG baseline (second row in Table \ref{tab:ablations}), considering spatial (S) coherence is key due to the amount of frames where the same object overlaps within a video sequence.
Combining spatial and a simple Motion (M) estimation, there is a small improvement in MOT17, and small decay for MOT20 and DanceTrack, probably due to the simplicity of the motion model.

On the other hand, temporal (T) coherence provides a considerable improvement in the pedestrian datasets, MOT17 and MOT20, and a small decay for DanceTrack. This is also due to the nature of the datasets as, in MOT17 and MOT20, pedestrians may leave the scene to never reappear, while in DanceTrack dancers are usually always present in the video, even if they are occluded for a long time.
We highlight the fact that MOT20 has a staggering increase in $31.5$ points in HOTA score when using spatial (S) and temporal (T) coherence during the clustering of tracklets. Weighting the object associations with spatio-temporal priors is specially needed in this dataset due to the high density of pedestrians in the scene, making the correct assignment difficult, combined with their small size due to camera positioning.

\smallskip
\noindent \textbf{Low fps setting}

\begin{figure}[h]
    \centering
    \includegraphics[width=\linewidth]{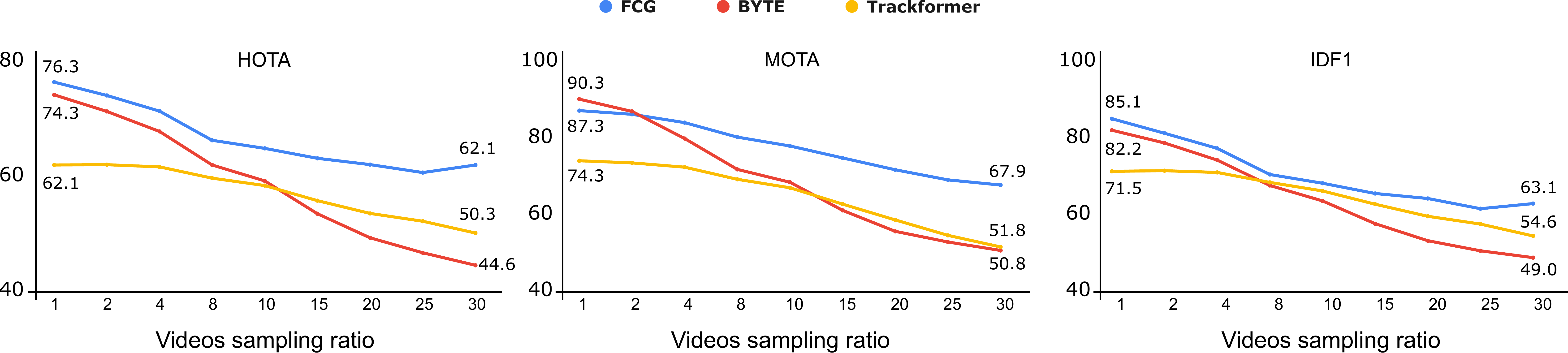}
    \caption{Low fps video setting experiment for state-of-the-art MOT trackers.}
    \label{fig:4_lowfps}
\end{figure}

A problem that current state-of-the-art trackers face comes when using them in low frames per second (fps) frameworks. We show this in Figure \ref{fig:4_lowfps}, where we tested three MOT trackers, BYTE \cite{zhang2022bytetrack}, Trackformer \cite{meinhardt2022trackformer}, and FCG -this work-, in the MOT17 \cite{MOT16} training set.
As the videos have different fps, we sub-sampled them by using a sampling ratio that ranges from $2$, corresponding to half the original video fps -e.g. from $30$ fps to $15$ fps-, to $30$, which corresponds to $1/30$ of the original video -e.g. from $30$ fps to $1$ fps-.

From the results, FCG seems to be more robust to lower fps settings than BYTE or Trackformer. With a performance of $62.1$ points of HOTA score on the most extreme sampling, $1$ frame every $30$ of the original video, FCG outperforms the other trackers, BYTE ($44.6$) and Trackformer ($50.3$). This comes as a side effect of the object association by using appearance features, and only using spatial and temporal priors as a weighting factor.

\begin{figure}[h]
    \centering
    \includegraphics[width=0.95\linewidth]{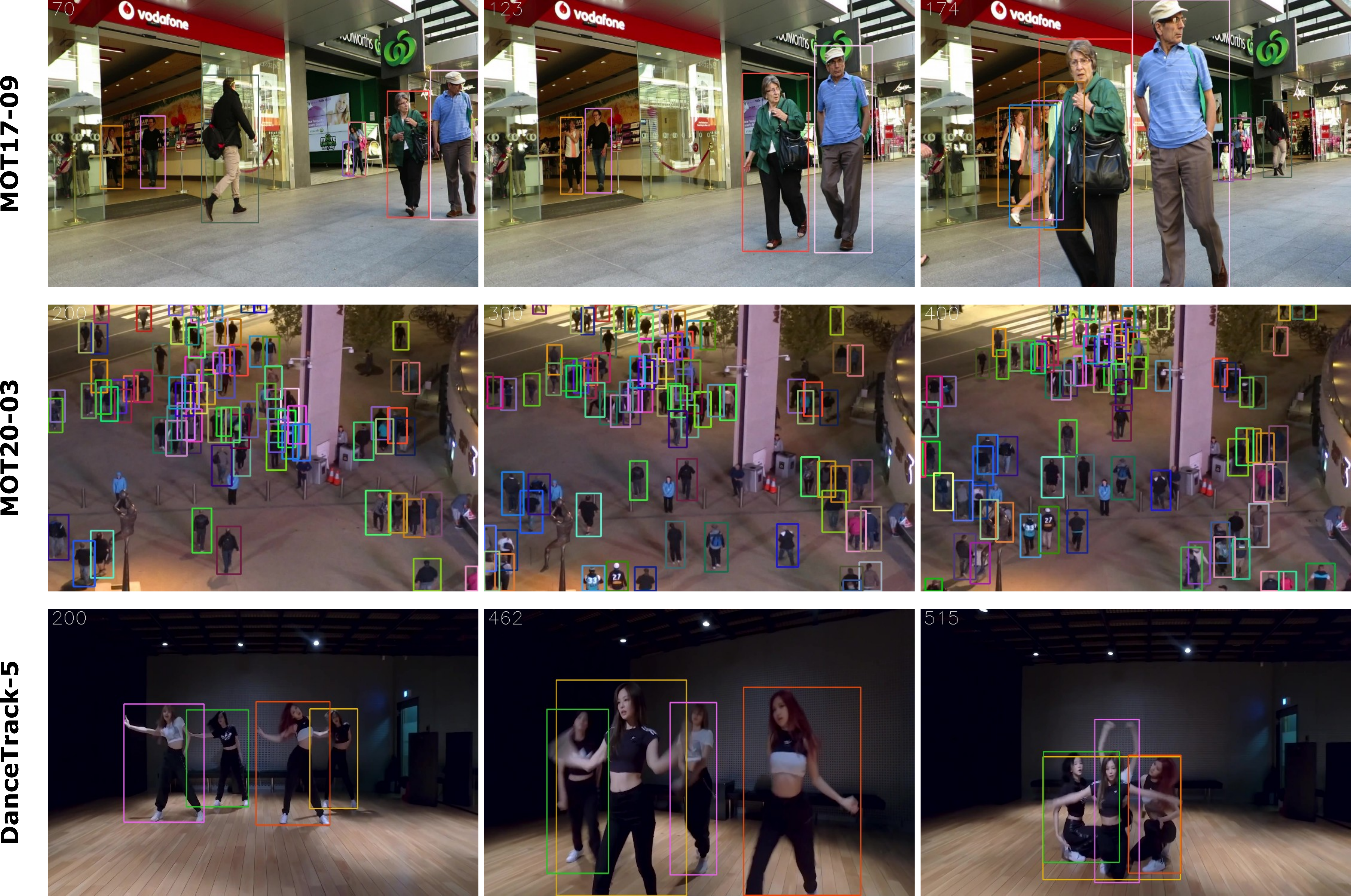}
    \caption{Qualitative results of FCG tracking for three sequences of the tested benchmarks, MOT17, MOT20 and DanceTrack. Same color corresponds to the same ID.}
    \label{fig:my_label}
\end{figure}

\section{Conclusions}

In this paper we explore multiple object tracking using object appearance as the principal element for the object association step, and present FCG, a MOT tracker that leverages on the idea that a specific object is similar in a temporal neighborhood to generate tracklets, and cluster them in a hierarchical way to form the final object tracks.
While being a much simpler method than other current MOT trackers, FCG achieves state-of-the-art performance on the DanceTrack benchmark, and presents competitive results in both MOT17 and MOT20 datasets.

\bibliography{bibliography.bib}

\begin{thebibliography}{54}
\providecommand{\natexlab}[1]{#1}
\providecommand{\url}[1]{\texttt{#1}}
\expandafter\ifx\csname urlstyle\endcsname\relax
  \providecommand{\doi}[1]{doi: #1}\else
  \providecommand{\doi}{doi: \begingroup \urlstyle{rm}\Url}\fi

\bibitem[Berclaz et~al.(2006)Berclaz, Fleuret, and Fua]{berclaz2006robust}
Jerome Berclaz, Francois Fleuret, and Pascal Fua.
\newblock Robust people tracking with global trajectory optimization.
\newblock In \emph{2006 IEEE Computer Society Conference on Computer Vision and
  Pattern Recognition (CVPR'06)}, volume~1, pages 744--750. IEEE, 2006.

\bibitem[Bergmann et~al.(2019)Bergmann, Meinhardt, and
  Leal-Taixe]{bergmann2019tracking}
Philipp Bergmann, Tim Meinhardt, and Laura Leal-Taixe.
\newblock Tracking without bells and whistles.
\newblock In \emph{Proceedings of the IEEE International Conference on Computer
  Vision}, pages 941--951, 2019.

\bibitem[Bernardin and Stiefelhagen(2008)]{bernardin2008evaluating}
Keni Bernardin and Rainer Stiefelhagen.
\newblock Evaluating multiple object tracking performance: the clear mot
  metrics.
\newblock \emph{EURASIP Journal on Image and Video Processing}, 2008:\penalty0
  1--10, 2008.

\bibitem[Bewley et~al.(2016)Bewley, Ge, Ott, Ramos, and
  Upcroft]{bewley2016simple}
Alex Bewley, Zongyuan Ge, Lionel Ott, Fabio Ramos, and Ben Upcroft.
\newblock Simple online and realtime tracking.
\newblock In \emph{2016 IEEE International Conference on Image Processing
  (ICIP)}, pages 3464--3468. IEEE, 2016.

\bibitem[Bras{\'o} and Leal-Taix{\'e}(2020)]{braso2020learning}
Guillem Bras{\'o} and Laura Leal-Taix{\'e}.
\newblock Learning a neural solver for multiple object tracking.
\newblock In \emph{Proceedings of the IEEE/CVF Conference on Computer Vision
  and Pattern Recognition}, pages 6247--6257, 2020.

\bibitem[Chen et~al.(2018)Chen, Ai, Zhuang, and Shang]{chen2018real}
Long Chen, Haizhou Ai, Zijie Zhuang, and Chong Shang.
\newblock Real-time multiple people tracking with deeply learned candidate
  selection and person re-identification.
\newblock In \emph{2018 IEEE international conference on multimedia and expo
  (ICME)}, pages 1--6. IEEE, 2018.

\bibitem[Chen et~al.(2019)Chen, Ai, Chen, and Zhuang]{chen2019aggregate}
Long Chen, Haizhou Ai, Rui Chen, and Zijie Zhuang.
\newblock Aggregate tracklet appearance features for multi-object tracking.
\newblock \emph{IEEE Signal Processing Letters}, 26\penalty0 (11):\penalty0
  1613--1617, 2019.

\bibitem[Chu and Ling(2019)]{chu2019famnet}
Peng Chu and Haibin Ling.
\newblock Famnet: Joint learning of feature, affinity and multi-dimensional
  assignment for online multiple object tracking.
\newblock In \emph{Proceedings of the IEEE/CVF International Conference on
  Computer Vision}, pages 6172--6181, 2019.

\bibitem[Dai et~al.(2021)Dai, Weng, Choi, Zhang, He, and Ding]{dai2021learning}
Peng Dai, Renliang Weng, Wongun Choi, Changshui Zhang, Zhangping He, and Wei
  Ding.
\newblock Learning a proposal classifier for multiple object tracking.
\newblock In \emph{Proceedings of the IEEE/CVF Conference on Computer Vision
  and Pattern Recognition}, pages 2443--2452, 2021.

\bibitem[Dendorfer et~al.(2020)Dendorfer, Rezatofighi, Milan, Shi, Cremers,
  Reid, Roth, Schindler, and Leal-Taix{\'e}]{dendorfer2020mot20}
Patrick Dendorfer, Hamid Rezatofighi, Anton Milan, Javen Shi, Daniel Cremers,
  Ian Reid, Stefan Roth, Konrad Schindler, and Laura Leal-Taix{\'e}.
\newblock Mot20: A benchmark for multi object tracking in crowded scenes.
\newblock \emph{arXiv preprint arXiv:2003.09003}, 2020.

\bibitem[Duan et~al.(2019)Duan, Bai, Xie, Qi, Huang, and
  Tian]{duan2019centernet}
Kaiwen Duan, Song Bai, Lingxi Xie, Honggang Qi, Qingming Huang, and Qi~Tian.
\newblock Centernet: Keypoint triplets for object detection.
\newblock In \emph{Proceedings of the IEEE International Conference on Computer
  Vision}, pages 6569--6578, 2019.

\bibitem[Feichtenhofer et~al.(2017)Feichtenhofer, Pinz, and
  Zisserman]{feichtenhofer2017detect}
Christoph Feichtenhofer, Axel Pinz, and Andrew Zisserman.
\newblock Detect to track and track to detect.
\newblock In \emph{Proceedings of the IEEE International Conference on Computer
  Vision}, pages 3038--3046, 2017.

\bibitem[Ge et~al.(2021)Ge, Liu, Wang, Li, and Sun]{yolox2021}
Zheng Ge, Songtao Liu, Feng Wang, Zeming Li, and Jian Sun.
\newblock Yolox: Exceeding yolo series in 2021.
\newblock \emph{arXiv preprint arXiv:2107.08430}, 2021.

\bibitem[He et~al.(2020)He, Liao, Liu, Liu, Cheng, and Mei]{he2020fastreid}
Lingxiao He, Xingyu Liao, Wu~Liu, Xinchen Liu, Peng Cheng, and Tao Mei.
\newblock Fastreid: A pytorch toolbox for general instance re-identification.
\newblock \emph{arXiv preprint arXiv:2006.02631}, 2020.

\bibitem[Hermans et~al.(2017)Hermans, Beyer, and Leibe]{hermans2017defense}
Alexander Hermans, Lucas Beyer, and Bastian Leibe.
\newblock In defense of the triplet loss for person re-identification.
\newblock \emph{arXiv preprint arXiv:1703.07737}, 2017.

\bibitem[Hornakova et~al.(2020)Hornakova, Henschel, Rosenhahn, and
  Swoboda]{hornakova2020lifted}
Andrea Hornakova, Roberto Henschel, Bodo Rosenhahn, and Paul Swoboda.
\newblock Lifted disjoint paths with application in multiple object tracking.
\newblock In \emph{International Conference on Machine Learning}, pages
  4364--4375. PMLR, 2020.

\bibitem[Huang et~al.(2008)Huang, Wu, and Nevatia]{huang2008robust}
Chang Huang, Bo~Wu, and Ramakant Nevatia.
\newblock Robust object tracking by hierarchical association of detection
  responses.
\newblock In \emph{European Conference on Computer Vision}, pages 788--801.
  Springer, 2008.

\bibitem[Lin et~al.(2014)Lin, Maire, Belongie, Hays, Perona, Ramanan,
  Doll{\'a}r, and Zitnick]{lin2014microsoft}
Tsung-Yi Lin, Michael Maire, Serge Belongie, James Hays, Pietro Perona, Deva
  Ramanan, Piotr Doll{\'a}r, and C~Lawrence Zitnick.
\newblock Microsoft coco: Common objects in context.
\newblock In \emph{European conference on computer vision}, pages 740--755.
  Springer, 2014.

\bibitem[Liu et~al.(2016)Liu, Anguelov, Erhan, Szegedy, Reed, Fu, and
  Berg]{liu2016ssd}
Wei Liu, Dragomir Anguelov, Dumitru Erhan, Christian Szegedy, Scott Reed,
  Cheng-Yang Fu, and Alexander~C Berg.
\newblock Ssd: Single shot multibox detector.
\newblock In \emph{European conference on computer vision}, pages 21--37.
  Springer, 2016.

\bibitem[Luiten et~al.(2020)Luiten, Osep, Dendorfer, Torr, Geiger, Leal-Taixe,
  and Leibe]{luiten2020hota}
Jonathon Luiten, Aljosa Osep, Patrick Dendorfer, Philip Torr, Andreas Geiger,
  Laura Leal-Taixe, and Bastian Leibe.
\newblock Hota: A higher order metric for evaluating multi-object tracking.
\newblock \emph{International Journal of Computer Vision}, 2020.

\bibitem[Luo et~al.(2019)Luo, Gu, Liao, Lai, and Jiang]{luo2019bag}
Hao Luo, Youzhi Gu, Xingyu Liao, Shenqi Lai, and Wei Jiang.
\newblock Bag of tricks and a strong baseline for deep person
  re-identification.
\newblock In \emph{Proceedings of the IEEE/CVF conference on computer vision
  and pattern recognition workshops}, pages 0--0, 2019.

\bibitem[Ma et~al.(2018)Ma, Tang, Black, and Van~Gool]{ma2018customized}
Liqian Ma, Siyu Tang, Michael~J Black, and Luc Van~Gool.
\newblock Customized multi-person tracker.
\newblock In \emph{Asian conference on computer vision}, pages 612--628.
  Springer, 2018.

\bibitem[Meinhardt et~al.(2022)Meinhardt, Kirillov, Leal-Taixe, and
  Feichtenhofer]{meinhardt2022trackformer}
Tim Meinhardt, Alexander Kirillov, Laura Leal-Taixe, and Christoph
  Feichtenhofer.
\newblock Trackformer: Multi-object tracking with transformers.
\newblock In \emph{Proceedings of the IEEE/CVF Conference on Computer Vision
  and Pattern Recognition}, pages 8844--8854, 2022.

\bibitem[Milan et~al.(2016)Milan, Leal-Taix\'{e}, Reid, Roth, and
  Schindler]{MOT16}
A.~Milan, L.~Leal-Taix\'{e}, I.~Reid, S.~Roth, and K.~Schindler.
\newblock {MOT}16: {A} benchmark for multi-object tracking.
\newblock \emph{arXiv:1603.00831 [cs]}, March 2016.
\newblock URL \url{http://arxiv.org/abs/1603.00831}.
\newblock arXiv: 1603.00831.

\bibitem[Pang et~al.(2021)Pang, Qiu, Li, Chen, Li, Darrell, and
  Yu]{pang2021quasi}
Jiangmiao Pang, Linlu Qiu, Xia Li, Haofeng Chen, Qi~Li, Trevor Darrell, and
  Fisher Yu.
\newblock Quasi-dense similarity learning for multiple object tracking.
\newblock In \emph{Proceedings of the IEEE/CVF conference on computer vision
  and pattern recognition}, pages 164--173, 2021.

\bibitem[Redmon and Farhadi(2017)]{redmon2017yolo9000}
Joseph Redmon and Ali Farhadi.
\newblock Yolo9000: better, faster, stronger.
\newblock In \emph{Proceedings of the IEEE conference on computer vision and
  pattern recognition}, pages 7263--7271, 2017.

\bibitem[Ren et~al.(2015)Ren, He, Girshick, and Sun]{ren2015faster}
Shaoqing Ren, Kaiming He, Ross Girshick, and Jian Sun.
\newblock Faster r-cnn: Towards real-time object detection with region proposal
  networks.
\newblock In \emph{Advances in neural information processing systems}, pages
  91--99, 2015.

\bibitem[Ristani and Tomasi(2018)]{ristani2018features}
Ergys Ristani and Carlo Tomasi.
\newblock Features for multi-target multi-camera tracking and
  re-identification.
\newblock In \emph{Proceedings of the IEEE conference on computer vision and
  pattern recognition}, pages 6036--6046, 2018.

\bibitem[Ristani et~al.(2016)Ristani, Solera, Zou, Cucchiara, and
  Tomasi]{ristani2016performance}
Ergys Ristani, Francesco Solera, Roger Zou, Rita Cucchiara, and Carlo Tomasi.
\newblock Performance measures and a data set for multi-target, multi-camera
  tracking.
\newblock In \emph{European Conference on Computer Vision}, pages 17--35.
  Springer, 2016.

\bibitem[Sneath and Sokal(1973)]{sneath1973unweighted}
Peter~HA Sneath and Robert~R Sokal.
\newblock Unweighted pair group method with arithmetic mean.
\newblock \emph{Numerical Taxonomy}, pages 230--234, 1973.

\bibitem[Stadler and Beyerer(2021{\natexlab{a}})]{stadler2021multi}
Daniel Stadler and J{\"u}rgen Beyerer.
\newblock Multi-pedestrian tracking with clusters.
\newblock In \emph{2021 17th IEEE International Conference on Advanced Video
  and Signal Based Surveillance (AVSS)}, pages 1--10. IEEE, 2021{\natexlab{a}}.

\bibitem[Stadler and Beyerer(2021{\natexlab{b}})]{stadler2021performance}
Daniel Stadler and J{\"u}rgen Beyerer.
\newblock On the performance of crowd-specific detectors in multi-pedestrian
  tracking.
\newblock In \emph{2021 17th IEEE International Conference on Advanced Video
  and Signal Based Surveillance (AVSS)}, pages 1--12. IEEE, 2021{\natexlab{b}}.

\bibitem[Stadler and Beyerer(2022)]{stadler2022modelling}
Daniel Stadler and J{\"u}rgen Beyerer.
\newblock Modelling ambiguous assignments for multi-person tracking in crowds.
\newblock In \emph{Proceedings of the IEEE/CVF Winter Conference on
  Applications of Computer Vision}, pages 133--142, 2022.

\bibitem[Sun et~al.(2020)Sun, Cao, Jiang, Zhang, Xie, Yuan, Wang, and
  Luo]{sun2020transtrack}
Peize Sun, Jinkun Cao, Yi~Jiang, Rufeng Zhang, Enze Xie, Zehuan Yuan, Changhu
  Wang, and Ping Luo.
\newblock Transtrack: Multiple object tracking with transformer.
\newblock \emph{arXiv preprint arXiv:2012.15460}, 2020.

\bibitem[Sun et~al.(2021)Sun, Cao, Jiang, Yuan, Bai, Kitani, and
  Luo]{peize2021dance}
Peize Sun, Jinkun Cao, Yi~Jiang, Zehuan Yuan, Song Bai, Kris Kitani, and Ping
  Luo.
\newblock Dancetrack: Multi-object tracking in uniform appearance and diverse
  motion.
\newblock \emph{arXiv preprint arXiv:2111.14690}, 2021.

\bibitem[Tokmakov et~al.(2021)Tokmakov, Li, Burgard, and
  Gaidon]{tokmakov2021learning}
Pavel Tokmakov, Jie Li, Wolfram Burgard, and Adrien Gaidon.
\newblock Learning to track with object permanence.
\newblock In \emph{Proceedings of the IEEE/CVF International Conference on
  Computer Vision}, pages 10860--10869, 2021.

\bibitem[Topkaya et~al.(2016)Topkaya, Erdogan, and
  Porikli]{topkaya2016tracklet}
Ibrahim~Saygin Topkaya, Hakan Erdogan, and Fatih Porikli.
\newblock Tracklet clustering for robust multiple object tracking using
  distance dependent chinese restaurant processes.
\newblock \emph{Signal, Image and Video Processing}, 10\penalty0 (5):\penalty0
  795--802, 2016.

\bibitem[Virtanen et~al.(2020)Virtanen, Gommers, Oliphant, Haberland, Reddy,
  Cournapeau, Burovski, Peterson, Weckesser, Bright, {van der Walt}, Brett,
  Wilson, Millman, Mayorov, Nelson, Jones, Kern, Larson, Carey, Polat, Feng,
  Moore, {VanderPlas}, Laxalde, Perktold, Cimrman, Henriksen, Quintero, Harris,
  Archibald, Ribeiro, Pedregosa, {van Mulbregt}, and {SciPy 1.0
  Contributors}]{2020SciPy-NMeth}
Pauli Virtanen, Ralf Gommers, Travis~E. Oliphant, Matt Haberland, Tyler Reddy,
  David Cournapeau, Evgeni Burovski, Pearu Peterson, Warren Weckesser, Jonathan
  Bright, St{\'e}fan~J. {van der Walt}, Matthew Brett, Joshua Wilson, K.~Jarrod
  Millman, Nikolay Mayorov, Andrew R.~J. Nelson, Eric Jones, Robert Kern, Eric
  Larson, C~J Carey, {\.I}lhan Polat, Yu~Feng, Eric~W. Moore, Jake
  {VanderPlas}, Denis Laxalde, Josef Perktold, Robert Cimrman, Ian Henriksen,
  E.~A. Quintero, Charles~R. Harris, Anne~M. Archibald, Ant{\^o}nio~H. Ribeiro,
  Fabian Pedregosa, Paul {van Mulbregt}, and {SciPy 1.0 Contributors}.
\newblock {{SciPy} 1.0: Fundamental Algorithms for Scientific Computing in
  Python}.
\newblock \emph{Nature Methods}, 17:\penalty0 261--272, 2020.
\newblock \doi{10.1038/s41592-019-0686-2}.

\bibitem[Wang et~al.(2019)Wang, Wang, Zhang, Gu, and Hwang]{wang2019exploit}
Gaoang Wang, Yizhou Wang, Haotian Zhang, Renshu Gu, and Jenq-Neng Hwang.
\newblock Exploit the connectivity: Multi-object tracking with trackletnet.
\newblock In \emph{Proceedings of the 27th ACM International Conference on
  Multimedia}, pages 482--490, 2019.

\bibitem[Wang et~al.(2018)Wang, Yuan, Chen, Li, and Zhou]{wang2018learning}
Guanshuo Wang, Yufeng Yuan, Xiong Chen, Jiwei Li, and Xi~Zhou.
\newblock Learning discriminative features with multiple granularities for
  person re-identification.
\newblock In \emph{Proceedings of the 26th ACM international conference on
  Multimedia}, pages 274--282, 2018.

\bibitem[Wang et~al.(2021{\natexlab{a}})Wang, Zheng, Pan, and
  Xu]{wang2021multiple}
Qiang Wang, Yun Zheng, Pan Pan, and Yinghui Xu.
\newblock Multiple object tracking with correlation learning.
\newblock In \emph{Proceedings of the IEEE/CVF Conference on Computer Vision
  and Pattern Recognition}, pages 3876--3886, 2021{\natexlab{a}}.

\bibitem[Wang et~al.(2021{\natexlab{b}})Wang, Sheng, Zhang, Wu, and
  Xiong]{wang2021general}
Shuai Wang, Hao Sheng, Yang Zhang, Yubin Wu, and Zhang Xiong.
\newblock A general recurrent tracking framework without real data.
\newblock In \emph{Proceedings of the IEEE/CVF International Conference on
  Computer Vision}, pages 13219--13228, 2021{\natexlab{b}}.

\bibitem[Wojke et~al.(2017)Wojke, Bewley, and Paulus]{wojke2017simple}
Nicolai Wojke, Alex Bewley, and Dietrich Paulus.
\newblock Simple online and realtime tracking with a deep association metric.
\newblock In \emph{2017 IEEE international conference on image processing
  (ICIP)}, pages 3645--3649. IEEE, 2017.

\bibitem[Wu et~al.(2021)Wu, Cao, Song, Wang, Yang, and Yuan]{wu2021track}
Jialian Wu, Jiale Cao, Liangchen Song, Yu~Wang, Ming Yang, and Junsong Yuan.
\newblock Track to detect and segment: An online multi-object tracker.
\newblock In \emph{Proceedings of the IEEE/CVF conference on computer vision
  and pattern recognition}, pages 12352--12361, 2021.

\bibitem[Wu et~al.(2018)Wu, Lin, Dong, Yan, Ouyang, and Yang]{wu2018exploit}
Yu~Wu, Yutian Lin, Xuanyi Dong, Yan Yan, Wanli Ouyang, and Yi~Yang.
\newblock Exploit the unknown gradually: One-shot video-based person
  re-identification by stepwise learning.
\newblock In \emph{Proceedings of the IEEE conference on computer vision and
  pattern recognition}, pages 5177--5186, 2018.

\bibitem[Yang et~al.(2021)Yang, Chang, Sakti, Wu, and Nakamura]{yang2021remot}
Fan Yang, Xin Chang, Sakriani Sakti, Yang Wu, and Satoshi Nakamura.
\newblock Remot: A model-agnostic refinement for multiple object tracking.
\newblock \emph{Image and Vision Computing}, 106:\penalty0 104091, 2021.

\bibitem[Yu et~al.(2021)Yu, Li, Han, and Wang]{yu2021relationtrack}
En~Yu, Zhuoling Li, Shoudong Han, and Hongwei Wang.
\newblock Relationtrack: Relation-aware multiple object tracking with decoupled
  representation.
\newblock \emph{arXiv preprint arXiv:2105.04322}, 2021.

\bibitem[Zeng et~al.(2021)Zeng, Dong, Wang, Zhang, and Wei]{zeng2021motr}
Fangao Zeng, Bin Dong, Tiancai Wang, Xiangyu Zhang, and Yichen Wei.
\newblock Motr: End-to-end multiple-object tracking with transformer.
\newblock \emph{arXiv preprint arXiv:2105.03247}, 2021.

\bibitem[Zhang et~al.(2020)Zhang, Sheng, Wu, Wang, Lyu, Ke, and
  Xiong]{zhang2020long}
Yang Zhang, Hao Sheng, Yubin Wu, Shuai Wang, Weifeng Lyu, Wei Ke, and Zhang
  Xiong.
\newblock Long-term tracking with deep tracklet association.
\newblock \emph{IEEE Transactions on Image Processing}, 29:\penalty0
  6694--6706, 2020.

\bibitem[Zhang et~al.(2021)Zhang, Wang, Wang, Zeng, and Liu]{zhang2021fairmot}
Yifu Zhang, Chunyu Wang, Xinggang Wang, Wenjun Zeng, and Wenyu Liu.
\newblock Fairmot: On the fairness of detection and re-identification in
  multiple object tracking.
\newblock \emph{International Journal of Computer Vision}, 129\penalty0
  (11):\penalty0 3069--3087, 2021.

\bibitem[Zhang et~al.(2022)Zhang, Sun, Jiang, Yu, Weng, Yuan, Luo, Liu, and
  Wang]{zhang2022bytetrack}
Yifu Zhang, Peize Sun, Yi~Jiang, Dongdong Yu, Fucheng Weng, Zehuan Yuan, Ping
  Luo, Wenyu Liu, and Xinggang Wang.
\newblock Bytetrack: Multi-object tracking by associating every detection box.
\newblock \emph{Proceedings of the European Conference on Computer Vision
  (ECCV)}, 2022.

\bibitem[Zhang et~al.(2017)Zhang, Wu, Zhang, and Zhang]{zhang2017multi}
Zhimeng Zhang, Jianan Wu, Xuan Zhang, and Chi Zhang.
\newblock Multi-target, multi-camera tracking by hierarchical clustering:
  Recent progress on dukemtmc project.
\newblock \emph{arXiv preprint arXiv:1712.09531}, 2017.

\bibitem[Zheng et~al.(2015)Zheng, Shen, Tian, Wang, Wang, and
  Tian]{zheng2015scalable}
Liang Zheng, Liyue Shen, Lu~Tian, Shengjin Wang, Jingdong Wang, and Qi~Tian.
\newblock Scalable person re-identification: A benchmark.
\newblock In \emph{Proceedings of the IEEE international conference on computer
  vision}, pages 1116--1124, 2015.

\bibitem[Zhou et~al.(2020)Zhou, Koltun, and
  Kr{\"a}henb{\"u}hl]{zhou2020tracking}
Xingyi Zhou, Vladlen Koltun, and Philipp Kr{\"a}henb{\"u}hl.
\newblock Tracking objects as points.
\newblock \emph{arXiv preprint arXiv:2004.01177}, 2020.

\end{thebibliography}
\end{document}